\documentclass[10pt,twocolumn]{article}

\usepackage[square,numbers]{natbib}
\usepackage{cvpr}
\usepackage{xcolor,xspace}
\usepackage{algorithm}
\usepackage[noend]{algpseudocode}
\usepackage{amsmath}
\usepackage{amssymb}
\usepackage{booktabs}
\usepackage{balance}
\usepackage{float}
\usepackage{amsfonts}      
\usepackage{url}

\newcommand{\comment}[1]{}

\usepackage{times}
\usepackage{epsfig}
\usepackage{graphicx}

\usepackage{subfigure}

\usepackage[pagebackref=true,breaklinks=true,colorlinks,bookmarks=false]{hyperref}

\cvprfinalcopy 

\def\cvprPaperID{XXXXX} 

\ifcvprfinal\fi
\begin{document}
\title{StepSaver: Predicting Minimum Denoising Steps for Diffusion Model Image Generation}
\author{Jean Yu\\
{\tt\small jean1.yu@intel.com}
\and
Haim Barad\\
{\tt\small haim.barad@intel.com}
}
\date{March 2024}
\maketitle
\thispagestyle{empty}

\begin{abstract}
In this paper, we introduce an innovative NLP model specifically fine-tuned to determine the minimal number of denoising steps required for any given text prompt. This advanced model serves as a real-time tool that recommends the ideal denoise steps for generating high-quality images efficiently. It is designed to work seamlessly with the Diffusion model, ensuring that images are produced with superior quality in the shortest possible time. Although our explanation focuses on the DDIM scheduler, the methodology is adaptable and can be applied to various other schedulers like Euler, Euler Ancestral, Heun, DPM2 Karras, UniPC, and more.  
This model allows our customers to conserve costly computing resources by executing the fewest necessary denoising steps to achieve optimal quality in the produced images.
\end{abstract}


\section{Introduction}
The focus in deep learning and computer vision has been on creating original, high-quality images. Although diffusion models produce images of remarkably high quality, several challenges, such as slow sampling because of numerous denoising steps, have needed to be overcome to make these models practical for image generation. The time it takes to generate an image from text using popular AI image-generating platforms can vary significantly based on several factors, including the complexity of the prompt. On a local setup with a high-performance GPU, Stable Diffusion can output images within seconds, usually between 10-30 seconds.  

\subsection{Diffusion Process}
As shown in Figure \ref{fig:diffusion_process} below, to develop a diffusion model, our goal is to train a model that can efficiently execute the noising and denoising processes. During the inference phase (when generating new images), we engage in the denoising process by starting with a noise vector and iteratively denoising it to produce a clear image. 

In the case of the stable diffusion model, it accepts textual input along with a seed. This textual input is processed through the CLIP model to create a textual embedding of dimensions 77x768, and the seed is utilized to generate Gaussian noise with dimensions 4x64x64, which serves as the initial latent image representation. Following this, the U-Net model denoises the random latent image representations, considering the text embeddings throughout the process. The U-Net outputs a predicted noise residual, which is employed to calculate conditioned latents using a scheduler algorithm. This cycle of denoising and text conditioning is repeated N times (by default, 50 iterations) to enhance the latent image representation. After completion of this sequence, the latent image representation (4x64x64) is transformed by the VAE decoder into the final output image (3x512x512).  

The scheduler algorithm plays a critical role in quality of generated images by managing how much noise is introduced to the input image at each stage of the denoising procedure, and then iteratively execute multiple steps to transform a noisy image into a clear one. The scheduler (preconfigured with N denoise steps) is also a performance bottleneck. The number of denoising steps determines the number of iterations the model takes to convert the noise to an image. It is a common belief that a higher number of steps generally leads to finer details and better image quality, but requires more computational resources and time. Currently, there are over ten different schedulers available for the denoising cycle. The flexibility of the Stable Diffusion model allows for easy modification, including swapping out schedulers.  

In summary, to address the slow-sampling issues of stable diffusion, the key is to find the optimal denoise steps. It is important to find those steps for each prompt in real time, so we do not add to the inference time. Ideally, this solution can handle all different schedulers available for the denoising cycle.

\subsection{Existing Solutions}
1. Existing platforms use fixed denoising steps for all prompts (typically 50 steps for the DDIM scheduler). This fixed setting can be very inefficient since GPU resources are wasted in cycles that no longer have an impact on the quality of generated images. More detailed results can be found in Section 7. 

2. Some platforms rely on users to experiment with various denoising steps (10, 20, 30, 50, ... 100, ...) and determine the steps based on examining the output images visually, but doing this for many prompts is not practical. Inexperienced users could use some help getting to the right range of denoising steps for their individual prompts. 

3. If the high usage of GPU resources is not a primary issue, one might consider opting for a greater fixed number of denoising steps, such as 100 steps. Theoretically, a higher number of steps can enhance the detail and clarity of images. However, it is important to note that increasing the number of steps does not necessarily guarantee improved quality in the generated images. 

4. We also considered a more obvious early exit approach, which requires modifying the scheduler code to interrupt the denoising loop before the predetermined steps. This approach requires significant changes to the algorithm. The essence of scheduler algorithms lies in their ability to gradually introduce and then methodically reduce random noise in data (according to a predetermined plan), thereby creating new data samples at the end of the process. Simply interrupting the scheduler, and any intermediate images will not have the same level of detail and definitions.  

\section{Details of StepSaver}
In this section we describe the method to define and quantify the association between denoise steps vs. quality/performance of image generation.

\subsection{Method to associate between denoise steps and image quality}
\subsubsection{Research Environment}
We acquired 2,322,632 images with 512x512 resolution, each accompanied by captions, from the LAION-Aesthetics v2 6+ subset. These captions serve as prompts for image generation. During evaluation, the original LAION-Aesthetic images are employed as reference or ground-truth images.  

Performance data is collected from experiments running on Habana Gaudi-1 devices. 

\subsubsection{Defining the Optimal Denoise Steps}
Figure \ref{fig:visual} displays six rows of generated images, each preceded by a related prompt. Within every row, the images are organized according to the number of denoising steps, starting from the lowest (10) and increasing to the highest (100) in increments of 10 steps. 
Skimming through the images in each row from left to right in Figure 7-1, there is a gradual and consistent improvement in the images from 10 to 30 steps. This improvement sometimes extends up to 50 steps. Beyond 60 steps, the images appear to change focus slightly. Although images produced with 100 steps can sometimes appear better, often using more than 50 steps does not lead to enhanced quality, and the performance significantly deteriorates. The impact on performance is explained in detail in the next section.  

\subsubsection{Optimal Denoise Steps Setting - Impact on Quality of Generated Images}
Examining the images generated for the prompt "Joel Robison The Black Dog" in Figure \ref{fig:deepdive} more closely, the model aims to create images of a dog in the style of photographer Joel Robison. From steps 10 to 50, we observe that the images develop consistently in a specific direction. However, from steps 60 to 80, the imagery unexpectedly shifts to depicting a house instead of an animal. Remarkably, by step 100, the model produces an image of a tree! In this case, I would say that 50 denoise steps produce an image of the highest quality. 

A similar pattern is observed in Figure \ref{fig:deepdive2} with the prompt 'Slow Cooker Salsa Chicken Tacos | via Midwest Nice Blog'. In the initial range of 10-30 steps, the model progressively refines elements such as avocado, tortillas, and cilantro. However, as we move from step 50 to 100, the model starts to introduce a lemon, but the tortillas begin to distort, and the cilantro leaves become so blurred that they are nearly indistinguishable. Once again, this indicates that a higher number of steps does not necessarily equate to better image quality. In this example, the best image can be generated in just 30 steps.  

\subsubsection{Optimal Denoise Steps Setting - Impact on Performance}
We ran experiments with Habana Gaudi1 and collected the average image generation time for hundreds of thousands of text prompts using the denoise step settings of 10 to 100. As shown in Figure \ref{fig:time-vs-steps}, the linear correlation between denoise steps and generation time is apparent. In other words, lower denoise steps directly reduce the generation time. 

\subsubsection{Calculating the Optimal Denoise Steps Based on SSIM Metric}
We first quantify how similar each pair of consecutive images is by calculating the Structural Similarity Index (SSIM) metric. The returned value is in the range of [0, 1]. A value of 0 indicates that the two given images are very different while a value of 1 indicates that the two given images are very similar or the same.

As the number of denoise steps increases in Figure \ref{fig:ssim}, from left to right the value of SSIM scores increases first from 0.5261 to 0.6762. The first decline in the sequence (from 0.6769 to 0.4103) suggests that the similarity between two consecutive images (from step 50 to step 60) is increasing, implying that the rate of improvement has slowed down at this point. We choose the first image of the two (in this case, 50) as the optimal denoise steps. 

We automated the calculation method described above and produced 287,340 denoise steps. Figure \ref{fig:data} shows examples of these steps along with the associated text prompts: 

Figure \ref{fig:data-summary} shows that the distribution of all 287,340 calculated denoise steps is not uniform, with concentrations specifically at 20, 30, and 50 steps. The cumulative totals for each of these denoising steps are 48,347, 162,783, and 76,210, respectively. 

\subsubsection{Metric for Measuring Image Quality}
Now that we can automate the process of finding the best number of denoising steps, how do we know that the quality of the images is better? The Frechet Inception Distance (FID) score is commonly used to measure the distance between the feature vectors of real and generated images. It serves as a gauge for assessing the quality of images produced by Diffusion Models, with lower scores indicating higher image quality.  

In Figure \ref{fig:quality-improvement}, the first 10 rows in the spreadsheet represent FID scores for output images generated with fixed denoising steps ranging from 10 to 100. The last row represents data from flexible denoise steps by calculation (described in the previous section). Data in the second to the last row (flexi-steps recommended by NLP) were generated from a new NLP model. We will explain the NLP model in the following sections. For now, it is sufficient to note that images generated with both flexi-steps (along with fixed 100-steps result in the highest quality images, as evidenced by the lowest FID scores. 

We use the FID for fixed 50 steps as the baseline because this is how the Stable Diffusion Inference pipeline is configured by default. The combo chart under the spreadsheet in Figure \ref{fig:quality-improvement} is the visual representation of how the denoise steps impact the FID scores relative to this baseline. Notably, the three bars to the right in the combo chart (associated with fixed 100, flexi-model recommended, and flexi-calculated, respectively) show the most significant quality enhancements. These three bars match the last three rows in the spreadsheet with the largest drop in FID scores from the baseline. 

\subsection{Denoise Steps Recommender Service (based on a new NLP model)}
In this section, we calculate the optimal steps for individual text prompts. As shown in Figure \ref{fig:distribution}, only 76,210 of the 287,340 prompts (approximately 26. 5\%) require the same number of steps as the baseline (50 steps).

73.5\% of the prompts only need 30 or 20 steps to achieve the best image quality. The quality of output images using lower and flexible steps (20, 30, or 50 steps) is significantly better than those using fixed denoise steps (including 50 to 100 steps). This is very encouraging.  

The remaining issue is that the optimal steps calculation takes longer than the image generation pipeline itself. Running the calculation during image generation defeats the purpose of speeding up Stable Diffusion inferences. As a result of our experiments and calculations so far, we have 287,340 rows of text prompts along with their calculated optimal denoise steps (Figure \ref{fig:data}). We will create our own denoise steps dataset and hope to train a new NLP model that provides denoise steps recommendations in real time. 

\subsubsection{Dataset Preprocessing for NLP Model Training}
After removing data rows with non-English characters. We observed that the distribution of the denoise steps is inherently unbalanced (as indicated in Figure \ref{fig:distribution}). To reduce model bias and ensure the accuracy of the new model, we decided to under-sample the majority class (30 steps), keeping 50-step rows unchanged, and ignore the 20-step class because it does not enough row count. By now, the dataset row count was reduced from 287,340 to 159,936 with an even distribution of two class labels: 30 and 50.  

This is not ideal because the 20-step class is not represented in the trained model. The aggregate quality of the images using the steps predicted by the NLP model will be slightly worse than the ones from direct calculation. The actual FID scores are provided in the last two rows in Figure \ref{fig:quality-improvement}.  

In addition, we reserved 2757 rows for testing. The remaining rows were split 9:1 between training and validation sets.  

As we continue to generate more prompts labeled with calculated steps (20, 30, 50), we will include the 20-step class, and model accuracy will be improved.  

\subsubsection{Model Training}
The following hyperparameters were used during the training: 

learning rate: 2e-03 

train batch size: 16 

eval batch size: 32 

num epochs: 5 

 Training achieves the following results on the evaluation set: 

Training Loss: 0.7269 

Validation BCE loss: 0.6973 

Test Accuracy: 0.6755 (0.737 with unbalanced dataset) 

Test F1: 0.68 (0.74 with unbalanced dataset) 

\subsubsection{Using Flexi-Steps Recommended by the NLP Model}
First, we use the new NLP model to predict the optimal denoise steps. Next, the Stable Diffusion pipeline takes the prompt and recommends denoise steps to generate images. Repeat the above for all 2757 prompts in the test dataset. An FID score is calculated for the resulting 2757 images against true images LAION-Aesthetics v2 6+ dataset. In Figure \ref{fig:quality-improvement}, the Flexi-Steps from recommendation is among the top 3 in terms of image quality (FID score).  

Figure \ref{fig:stepsaver_performance} shows a summary of the total number of recommended prompts for each class label.
Below is a performance and quality summary comparing fixed-50-steps, fixed-100-steps, and flexible-recommended-steps. The overall image generation time refers to the average image generation time per image in Figure \ref{fig:time-vs-steps}: 

Overall generation time 

50-fixed steps: 3.72*2757 = 10,256.04 seconds (2.85 hours) 

100-fixed steps: 7.36*2757 = 20,291.52 seconds (5.64 hours) 

Flexible-recommended-steps: 2337*2.25 + 420*3.72 = 6820.65 seconds (1.89 hours)

\begin{figure*}
  \centering
  \includegraphics[width=1\linewidth]{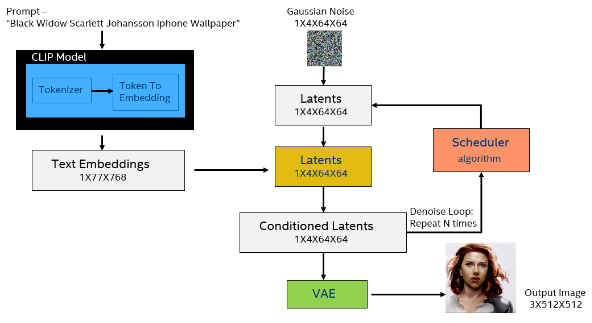}
  \caption{The Diffusion Process}
\label{fig:diffusion_process}
\end{figure*}

\begin{figure*}
  \centering
  \includegraphics[width=1\linewidth]{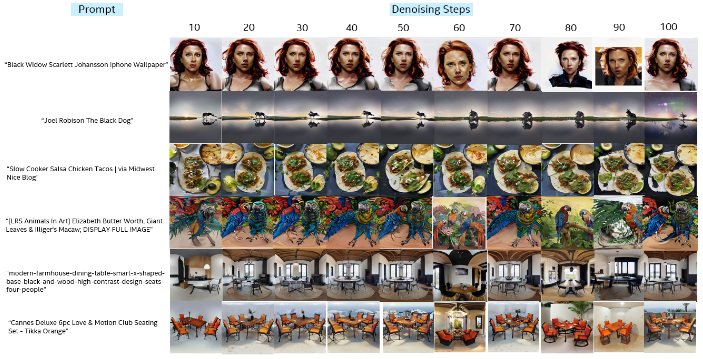}
  \caption{visual display of generated images from 6 prompts and why denoise steps matter}
\label{fig:visual}
\end{figure*}

\begin{figure*}
  \centering
  \includegraphics[width=1\linewidth]{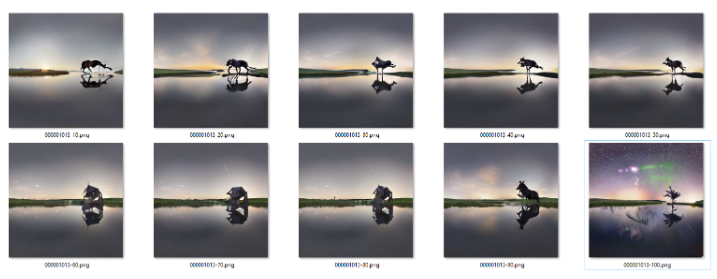}
  \caption{Deep Dive – Generated image for prompt “Joel Robison The Black Dog”}
\label{fig:deepdive}
\end{figure*}

\begin{figure*}
  \centering
  \includegraphics[width=1\linewidth]{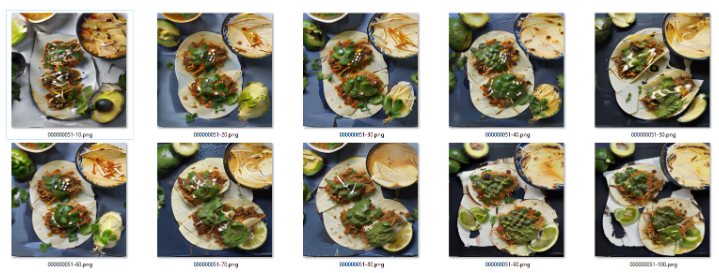}
  \caption{Deep Dive – Generated images for prompt “Slow Cooker Salsa Chicken Tacos | via Midwest Nice Blog”}
\label{fig:deepdive2}
\end{figure*}

\begin{figure*}
  \centering
  \includegraphics[width=1\linewidth]{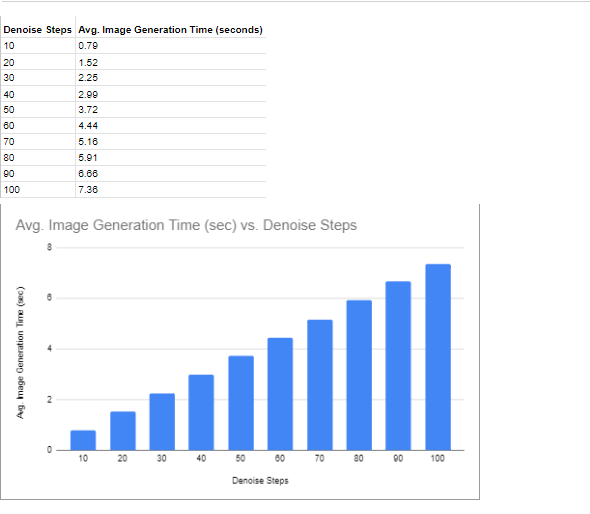}
  \caption{Average Image Generation Time vs Denoise Steps (Accelerator: Habana Gaudi1)}
\label{fig:time-vs-steps}
\end{figure*}

\begin{figure*}
  \centering
  \includegraphics[width=1\linewidth]{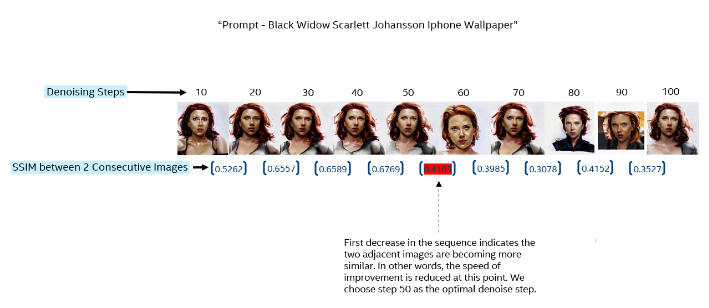}
  \caption{SSIM metric and definition of optimal denoise steps}
\label{fig:ssim}
\end{figure*}

\begin{figure*}
  \centering
  \includegraphics[width=1\linewidth]{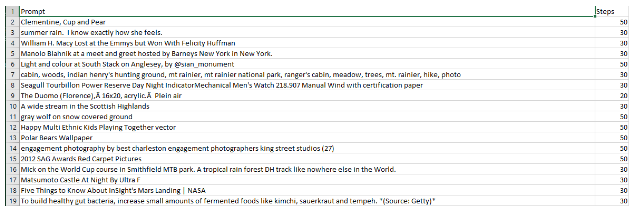}
  \caption{Optimal Denoise Steps by Calculation}
\label{fig:data}
\end{figure*}

\begin{figure*}
  \centering
  \includegraphics[width=1\linewidth]{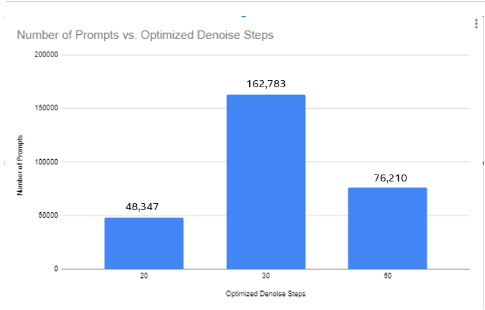}
  \caption{Distribution of Optimal Denoise Steps (by calculation)}
\label{fig:distribution}
\end{figure*}

\begin{figure*}
  \centering
  \includegraphics[width=1\linewidth]{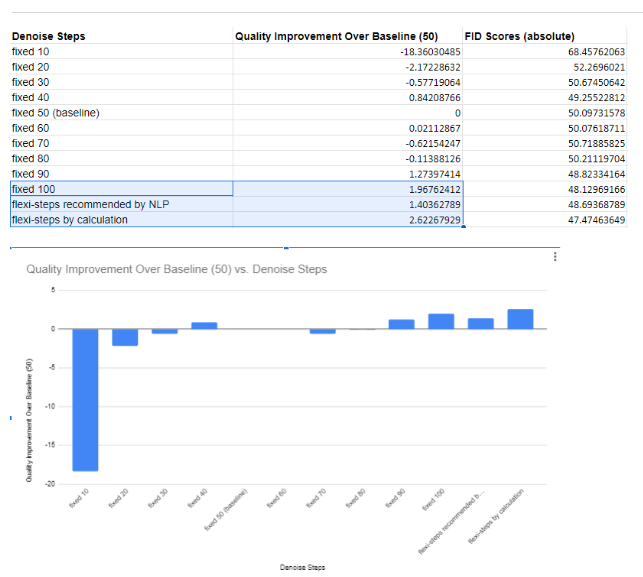}
  \caption{FID Scores - Image Quality Measurement (lower FID means higher quality)}
\label{fig:quality-improvement}
\end{figure*}

\begin{figure*}
  \centering
  \includegraphics[width=1\linewidth]{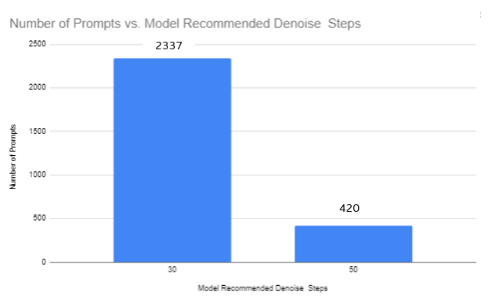}
  \caption{Number of Prompts vs. Model Recommended Denoise Steps}
\label{fig:data-summary}
\end{figure*}

\begin{figure*}
  \centering
  \includegraphics[width=1\linewidth]{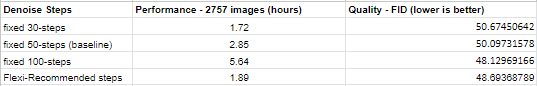}
  \caption{StepSaver Performance}
\label{fig:stepsaver_performance}
\end{figure*}

\clearpage
{\small
\bibliographystyle{ieeetr}
\nocite{*}
\bibliography{references}
}

\end{document}